\documentclass[sigconf]{acmart}

\AtBeginDocument{%
  }

\copyrightyear{2025}
\acmYear{2025}
\setcopyright{acmlicensed}
\acmConference[KDD '25] {Proceedings of the 31st ACM SIGKDD Conference on Knowledge Discovery and Data Mining V.1}{August 3--7, 2025}{Toronto, ON, Canada.}
\acmBooktitle{Proceedings of the 31st ACM SIGKDD Conference on Knowledge Discovery and Data Mining V.1 (KDD '25), August 3--7, 2025, Toronto, ON, Canada}
\acmISBN{979-8-4007-1245-6/25/08}
\acmDOI{10.1145/3690624.3709166}

\usepackage{graphicx}
\usepackage{amsthm}
\usepackage{amsmath}
\usepackage{mathrsfs}
\usepackage{booktabs}
\usepackage[linesnumbered,ruled,vlined]{algorithm2e}
\usepackage{multirow}
\usepackage{float}
\usepackage{makecell}
\usepackage{algpseudocode}

\newcommand{\TheName}{\texttt{PAI}}
\newif\ifshownotes
\shownotestrue

\newcommand{\lt}[2][]{%
  \ifshownotes%
    \ifx&#1&%
      \textbf{\textcolor{orange}{[Liantao: #2]}}%
    \else%
      {\textcolor{lightgray}{[(#1) Liantao: #2]}}%
    \fi%
  \fi%
}

\begin{document}

\title[Learnable Prompt as Pseudo-Imputation]{Learnable Prompt as Pseudo-Imputation: Rethinking the Necessity of Traditional EHR Data Imputation in Downstream Clinical Prediction}


\author{Weibin Liao}
\orcid{0000-0002-9682-9934}
\affiliation{
    \institution{Peking University}
    \department{School of Computer Science}
    \city{Beijing}
    \country{China}}
\affiliation{
    \institution{Ministry of Education}
    \department{Key Laboratory of High Confidence Software Technologies}
    \city{Beijing}
    \country{China}}
\email{liaoweibin@stu.pku.edu.cn}

\author{Yinghao Zhu}
\orcid{0000-0002-2640-6477}
\affiliation{
    \institution{Peking University}
    \department{National Engineering Research Center for Software Engineering}
    \city{Beijing}
    \country{China}}

\author{Zhongji Zhang}
\orcid{0009-0006-5454-6764}
\author{Yuhang Wang}
\orcid{0009-0003-0141-3827}
\author{Zixiang Wang}
\orcid{0009-0000-1257-9580}
\affiliation{
    \institution{Peking University}
    \department{School of Software and Microelectronics}
    \city{Beijing}
    \country{China}}
\affiliation{
    \institution{Ministry of Education}
    \department{Key Laboratory of High Confidence Software Technologies}
    \city{Beijing}
    \country{China}}



\author{Xu Chu}
\orcid{0000-0002-0520-7196}
\affiliation{
    \institution{Peking University}
    \department[0]{School of Computer Science}
    \department[1]{Center on Frontiers of Computing Studies}
    \city{Beijing}
    \country{China}}
\affiliation{
    \institution{Ministry of Education}
    \department{Key Laboratory of High Confidence Software Technologies}
    \city{Beijing}
    \country{China}}
\affiliation{
    \institution{Peking University}
    \department{Peking University Information Technology Institute (Tianjin Binhai)}
    \city{Tianjin}
    \country{China}}

\author{Yasha Wang}
\orcid{0000-0002-8026-9688}
\affiliation{
    \institution{Peking University}
    \department{National Engineering Research Center for Software Engineering}
    \city{Beijing}
    \country{China}}
\affiliation{
    \institution{Ministry of Education}
    \department{Key Laboratory of High Confidence Software Technologies}
    \city{Beijing}
    \country{China}}
\email{wangyasha@pku.edu.cn}

\author{Liantao Ma}\authornote{Corresponding Author.}
\orcid{0000-0001-5233-0624}
\affiliation{
    \institution{Peking University}
    \department{National Engineering Research Center for Software Engineering}
    \city{Beijing}
    \country{China}}
\affiliation{
    \institution{Ministry of Education}
    \department{Key Laboratory of High Confidence Software Technologies}
    \city{Beijing}
    \country{China}}
\email{malt@pku.edu.cn}

\renewcommand{\shortauthors}{Weibin Liao et al.}


\begin{abstract}
Analyzing the health status of patients based on Electronic Health Records (EHR) is a fundamental research problem in medical informatics. 
The presence of extensive missing values in EHR makes it challenging for deep neural networks (DNNs) to directly model the patient's health status.
Existing DNNs training protocols, including \textit{Impute-then-Regress Procedure} and \textit{Jointly Optimizing of Impute-n-Regress Procedure}, require the additional imputation models to reconstruction missing values.
However, \textit{Impute-then-Regress Procedure} introduces the risk of injecting imputed, non-real data into downstream clinical prediction tasks, resulting in power loss, biased estimation, and poorly performing models, while \textit{Jointly Optimizing of Impute-n-Regress Procedure} is also difficult to generalize due to the complex optimization space and demanding data requirements.
Inspired by the recent advanced literature of learnable prompt in the fields of NLP and CV, in this work, we rethought the necessity of the imputation model in downstream clinical tasks, and proposed Learnable \underline{P}rompt \underline{a}s Pseudo-\underline{I}mputation (\TheName{}) as a new training protocol to assist EHR analysis.
\TheName{} no longer introduces any imputed data but constructs a learnable prompt to model the implicit preferences of the downstream model for missing values, resulting in a significant performance improvement for \textit{all state-of-the-arts EHR analysis models} on four real-world datasets across two clinical prediction tasks.
Further experimental analysis indicates that \TheName{} exhibits higher robustness in situations of data insufficiency and high missing rates. 
More importantly, as a plug-and-play protocol, \TheName{} can be easily integrated into any existing or even imperceptible future EHR analysis models.
The code of this work is deployed publicly available at \url{https://github.com/MrBlankness/PAI} to help the research community reproduce the results and assist the EHR analysis tasks.

\end{abstract}




\maketitle

\section{Introduction}

\begin{figure*}[t]
\centering
\includegraphics[width=0.98\textwidth]{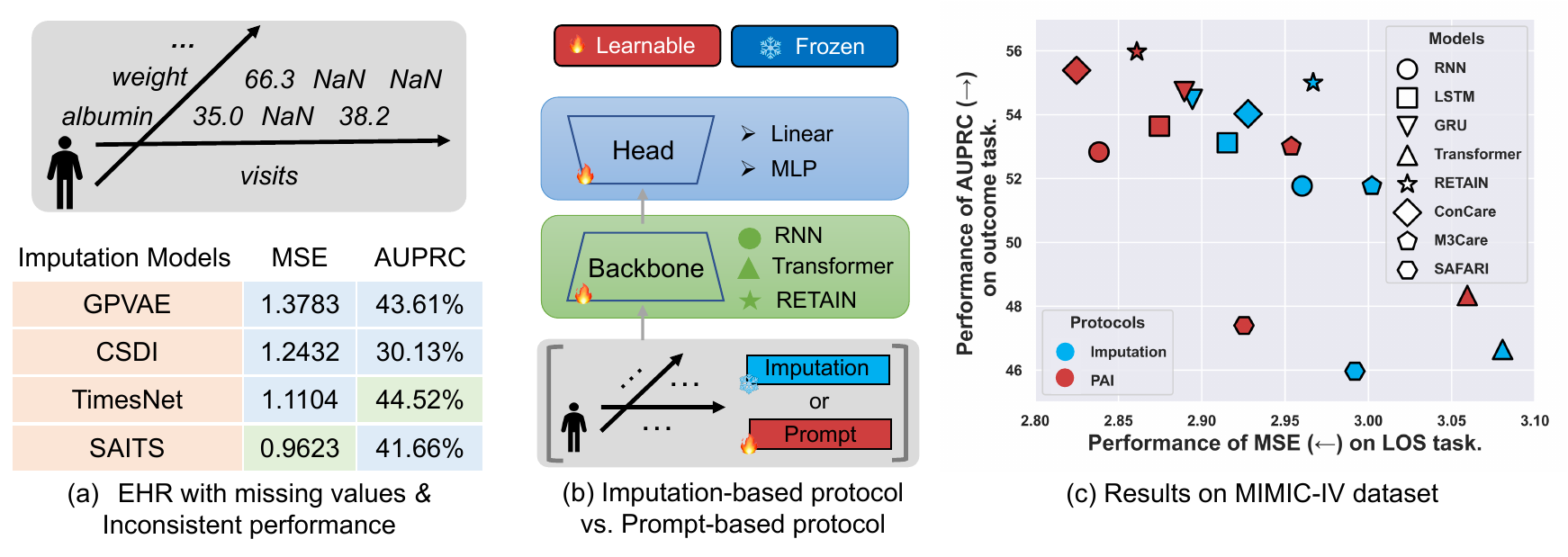}
\caption{Imputation-based protocol \textit{vs.} Prompt-based protocol. 
(a) illustrates typical examples of EHR with missing values and the inconsistency in performance between existing imputation models on the EHR imputation task and downstream outcome prediction tasks.
(b) presents a typical framework of EHR analysis models and highlights the differences between the imputation-based protocol and the prompt-based protocol.
(c) illustrates that on the MIMIC-IV dataset, compared to the imputation-based protocol, the prompt-based protocol improves the performance of all the baseline backbone models.
} \label{fig:overview}
\end{figure*}

Electronic Health Records (EHR), serving as a form of multivariate time series data~\cite{ma2023aicare,gao2024comprehensive,ma2023mortality}, play a crucial role in various aspects of healthcare, including clinical decision-making, disease modeling, and outcome prediction~\cite{shickel2017deep}. 
In recent years, Deep Neural Networks (DNNs)~\cite{10189221,10135109} have exhibited state-of-the-art performance in extracting valuable insights from the extensive and multifaceted EHR datasets~\cite{zhang2022m3care,ma2022safari,zhu2023leveraging}. 
However, EHR data often suffers from issues of incompleteness~\cite{ma2020concare} (as shown in Fig.~\ref{fig:overview}(a)) for intentional (e.g., when specific tests are unnecessary for the patient) or unintentional (e.g., due to a lack of routine checkups or follow-ups) reasons, making it challenging for DNNs, models relying on structured input, to directly model the patient’s health status.
Thus, overcoming the issue of data incompleteness in EHR remains a fundamental question and a significant area of focus in ongoing research within the healthcare informatics community.

Existing deep learning training protocols mandate researchers to address missing values in EHR through imputation methods before utilizing DNNs for patient health diagnostics. 
These \textit{imputation-based training protocols} can be divided into the following two types, namely the \textit{Impute-then-Regress Procedure} and \textit{Jointly Optimizing of Impute-n-Regress Procedure}.
The \textit{Impute-then-Regress Procedure} employs imputation serves as a preliminary step for clinical prediction, using methods such as mean substitution, regression, hot deck~\cite{madow1983incomplete}, tree-based~\cite{doove2014recursive}, as well as advanced statistical methods~\cite{marcoulides2013advanced,yoshikawa2020functional}.
However, on one hand, \textit{Impute-then-Regress Procedure} introduces the risk of injecting imputed, non-authentic data into downstream clinical prediction tasks, leading to a power loss, biased estimates~\cite{ayilara2019impact,van2020rebutting}, and underperforming models~\cite{li2021imputation}.
On the other hand, some recently proposed typical methods used in \textit{Impute-then-Regress Procedure}, including GPVAE~\cite{fortuin2020gpvae}, CSDI~\cite{tashiro2021csdi}, TimesNet~\cite{wu2022timesnet}, and SAITS~\cite{du2023saits}, have achieved inconsistent performance in imputation tasks and clinical prediction tasks as shown in Fig.~\ref{fig:overview}(a). From Fig.~\ref{fig:overview}(a) we found that SAITS~\cite{du2023saits} achieves optimal performance in imputation tasks, but the imputed values does not perform as well in downstream prediction tasks compared to TimesNet~\cite{wu2022timesnet}.
Therefore, we reasonably infer that \textit{the conditional expectation of the missing values observed by the preliminary imputation model is not the optimal target for the prediction tasks}.

Some recent works~\cite{le2021sa} have arrived at similar observations and have established theoretical arguments for this, leading to the development of methods for \textit{Jointly Optimizing of Impute-n-Regress Procedure}, which adopt a basic strategy of coupling the imputation model and the prediction model, performing joint optimization in the same training process. 
Although proven effective, \textit{Jointly Optimizing of Impute-n-Regress Procedure} implies a more complex optimization space. Furthermore, some existing works~\cite{miao2021generative,weng2024joint} require the injection of \textit{complete EHR} data to supervise the parameter optimization of the imputation model during joint learning, which is often not available in real scenarios~\cite{groenwold2020informative,tan2023informative}. 
Therefore, only a few works~\cite{miao2021generative,weng2024joint} have focused on this area, and most recent works~\cite{zhang2022m3care,ma2022safari} still follow the \textit{Impute-then-Regress Procedure}. 
In summary, \textit{an elegant, unsupervised, plug-and-play, and highly scalable strategy for handling missing EHR data is still worth further exploration}.

Recently, prompt learning has attracted extensive research and drawn significant attention in the domains of natural language processing (NLP) and computer vision (CV).
Unlike traditional definitions, the prompt is no longer considered a set of handcrafted instructional texts but is viewed as a set of continuously optimizable continuous vectors~\cite{liu2021p,li2021prefix,lester2021power} embedded in the input space of the language model.
This aligns with our previously mentioned hypothesis that handcrafted or prior model-constructed instructional texts (imputed values) are not the optimal target for downstream language models (EHR analysis models).
Inspired by this, we aim to \textit{eliminate the need for imputation models essential in traditional training protocols. As an alternative, we construct a continuous vector during downstream EHR analysis and optimize it in the process, allowing DNNs to learn solely based on real data without relying on manual or automatic data imputation as required in traditional protocols.}

In this work, we propose a new training protocol called Learnable \underline{P}rompt \underline{a}s Pseudo-\underline{I}mputation (\TheName{}) for EHR analysis without imputation. 
Specifically, \TheName{} trains DNNs with a learnable prompt vector. The prompt vector can be initialized with either random values or statistical information.
During training, we simply update the prompt vectors using the training loss of the downstream analysis task, without introducing additional losses to enhance model complexity. The gradients can be back-propagated all the way through the DNNs, capturing parameter preferences related to missing values in the model for learning prompts.
We claim that despite being a new protocol, \textit{\TheName{} does not replace any parameterized components of the original protocol (see shown in Fig.~\ref{fig:overview}(b)). Therefore, any existing model for EHR analysis tasks under the original protocol can seamlessly adopt the new protocol without incurring additional costs and gain further performance improvements (see shown in Fig.~\ref{fig:overview}(c))}. 

To demonstrate the effectiveness of \TheName{}, we conducted benchmark tests on various EHR analysis models, various datasets and various tasks, to verify the effectiveness of \TheName{} in various EHR analysis scenarios.
Experimental results indicate that \textit{\TheName{} significantly enhances the performance of 8 models across 2 EHR analysis tasks on 4 datasets, showcasing substantial advantages over traditional imputation-based protocol}. 
Furthermore, \TheName{} exhibits stronger robustness on datasets with smaller training sizes and higher rates of data missing. 
In summary, we make the following contributions:
\begin{enumerate}
    \item \textbf{Insightly}, we reassess the necessity of imputation model in traditional EHR analysis protocol and proposed a new protocol. To the best of our knowledge, this is the first framework revealing superior performance without imputation than methods with imputation.
    \item \textbf{Methodologically}, to address the dependence of DNNs on structured input, we propose a plug-and-play approach based on continuous prompt learning, identify common patterns in the prompt for EHR analysis and directly improved the performance of downstream tasks by continuously optimizing the prompt. As a plug-and-play method without imputation of raw data or modifying backbone models, \TheName{} is easy to follow and thus can be seamlessly integrated into any existing DNNs analysis frameworks.
    \item \textbf{Experimentally}, we evaluate \TheName{} on four real-world datasets, demonstrating that the proposed protocol optimizes the performance of \textit{all} EHR analysis models in clinical prediction tasks and surpasses existing imputation-based protocol. Extensive experiments further demonstrate the outstanding performance of \TheName{} in model generalization.
\end{enumerate}

\section{Related Works}

\subsection{Imputation for EHR}

\paragraph{Impute-then-Regress Procedure}\label{sec:I-t-R}

The \textit{Impute-then-Regress Procedure} requires the imputation of missing values in EHR data to be accomplished in a pre-processing manner, and the static imputed EHR data is used in downstream prediction tasks.
There is a large body of literature on the imputation of missing values in EHR, and we only mention a few closely related ones.
Some statistical methods for imputing time series data are generally straightforward and involve substituting missing values with basic statistical measures, such as zero, the mean value, or the last observed value \cite{wells2013strategies}. 
Recently, advancements in deep learning have introduced more sophisticated approaches for imputing time series data. These approaches include the use of recurrent neural networks (RNNs) \cite{cao2018brits}, deep generative models \cite{liu2019naomi,luo2019e2gan}, variational autoencoders (VAEs) \cite{fortuin2020gpvae,ramchandran2021longitudinal}, and self-attention-based models \cite{bansal2021missing,shan2023nrtsi}. 

While these deep learning models have achieved state-of-the-art performance in the task of imputation, they are typically not directly applied to downstream EHR analysis tasks.
The primary challenge with these imputation methods lies in the instability of the imputed values. This instability can lead to several issues, including power loss and biased estimates \cite{ayilara2019impact,van2020rebutting}, which in turn result in underperforming models \cite{li2021imputation}. These challenges underscore the limitations of relying on imputation for downstream analysis tasks.
In light of these challenges, \textit{we propose an innovative approach that focuses on directly optimizing the downstream task without resorting to imputation.} By bypassing the imputation process, we aim to avoid the amplification of intermediate errors that often arise from imputed data. This strategy not only simplifies the analytical process but also enhances the overall robustness and accuracy of the downstream EHR analysis.

\paragraph{Jointly Optimizing of Impute-n-Regress Procedure}\label{sec:Jo-I-a-R}

The most relevant works to our study is the \textit{Jointly Optimizing of Impute-n-Regress Procedure}. These works~\cite{miao2021generative,weng2024joint}, like us, believe that the conditional expectation of the observed missing values by the imputation model is not the best goal for the downstream prediction task. To overcome this problem, the \textit{Jointly Optimizing of Impute-n-Regress Procedure} establishes a complementary relationship by combining imputation and prediction to obtain an unbiased estimate of the original data in the downstream task objective.

To the best of our knowledge, only very few works have been dedicated to verifying the effectiveness of the \textit{Jointly Optimizing of Impute-n-Regress Procedure} on EHR data. 
Among them, SSGAN~\cite{miao2021generative} proposes a semi-supervised (for imputation) framework that combines the generative adversarial network (GAN) for imputation with the RNN for prediction, enabling the imputation model to learn the label information of the classification task. 
Recently, similar to SSGAN. MVIIL-GAN~\cite{weng2024joint} combines GAN with the classification model, overcomes the challenges of imbalance and incompleteness of EHR data and is optimized under a fully supervised (for imputation) framework. 
Although proven to be effective, both SSGAN and MVIIL-GAN need to introduce the supervision signal of the imputation task and have to seek \textbf{complete EHR data}, and introduce GAN with a significant number of extra parameters, severely affecting the generalization of this type of method.
In contrast, \textit{\TheName{}, as an unsupervised, plug-and-play method, can directly enhance the performance of any EHR analysis model end-to-end.}

\subsection{Prompt Learning}\label{ssec:prompt-learning}

\paragraph{Prompt for Natural Language Processing}

This subject originates from the field of natural language processing (NLP). The goal is to leverage pre-trained language models, such as BERT~\cite{devlin2018bert} and GPT~\cite{radford2019language}, as knowledge bases to extract valuable information for downstream tasks~\cite{petroni2019language,liao2024tpo}. This process is often framed as a ``fill-in-the-blank" cloze test. For example, in sentiment classification, the model might be tasked with predicting the masked token in the sentence ``No reason to watch. It was [MASK]" as either ``positive" or ``negative". The challenge lies in designing prompts (templates) that the model can easily understand.

Instead of manually creating these prompts, research in prompt learning aims to automate the process using manageable amounts of labeled data. LPAQA~\cite{jiang2020can} employs text mining and paraphrasing to generate a set of candidate prompts, selecting the most optimal ones based on training accuracy. AutoPrompt~\cite{shin2020autoprompt} introduces a gradient-based approach that selects the best tokens from a vocabulary to maximize changes in gradients according to label likelihood. Our research aligns closely with continuous prompt learning methods~\cite{lester2021power,li2021prefix,zhong2021factual}, which focus on transforming a prompt into a set of continuous vectors that can be optimized end-to-end with respect to an objective function.

\paragraph{Prompt for EHR}

In the realm of EHR research, prompt learning is an emerging direction that has only recently begun to be explored~\cite{wang2022promptehr,cui2023medtem2,taylor2023clinical}. These pioneering efforts primarily focus on leveraging language models to enhance or generate EHR data. For example, the promptEHR framework~\cite{wang2022promptehr} conceptualizes EHR generation as a text-to-text translation task utilizing language models, which facilitates highly flexible event imputation during the data generation process.

Unlike the above-mentioned works that limit prompt as a textual representation, our inspiration is consistent with the contribution of prompt in the recent NLP fields. Workers in the NLP fields attempt to use the method of directly optimizing learnable vectors in downstream tasks to avoid the manual construction of prompt text. Similar to this inspiration, we aim to \textit{apply this learnable vector in the scenario of EHR analysis tasks with missing values, thereby avoiding the modeling of missing values by manual or imputation models.}

\section{Methodology}\label{sec:methodology}
\begin{figure}[!t]
\centering
\includegraphics[width=0.48\textwidth]{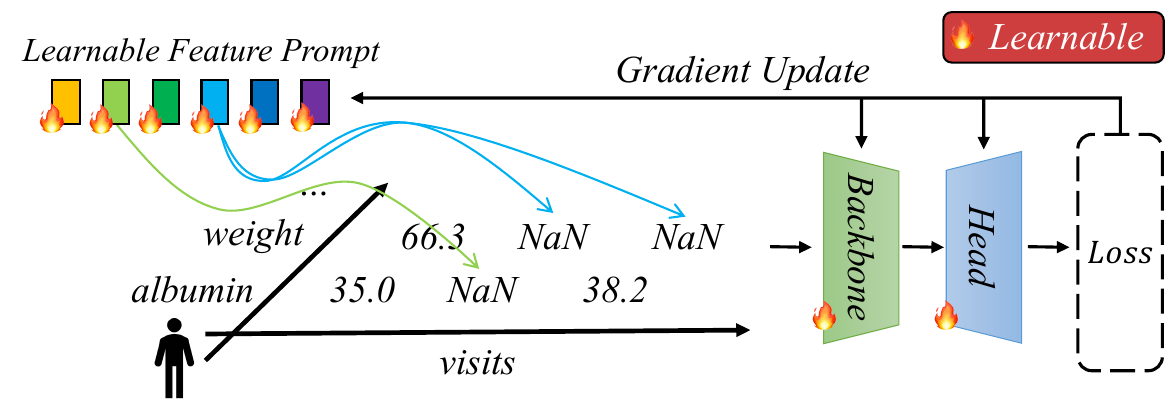}
\caption{The framework of \TheName{}. The main idea is to model missing values using a learnable prompt vector, which can be optimized by minimizing the loss of downstream tasks.} \label{fig:framework}
\end{figure}

We propose Learnable \underline{P}rompt \underline{a}s Pseudo-\underline{I}mputation (\TheName{}) for training DNNs on EHR with missing values. \TheName{} introduces a learnable prompt to model the parameter preferences of DNNs for missing values in EHR, rather than structuring data with imputation. The overall framework is presented in Fig.~\ref{fig:framework}. We first define the notations in Sec.~\ref{ssec:preliminaries}, then describe \TheName{} formally in Sec.~\ref{ssec:pai}. To aid reader comprehension, please note that in the subsequent manuscript, ``prompt'' refers to a set of \textbf{learnable feature vectors} rather than traditional text instructions (see Section.~\ref{ssec:prompt-learning} for details).

\subsection{Preliminaries}\label{ssec:preliminaries}
\paragraph{Multivariate time series}
We consider $N$ multivariate time series with missing values. Let us denote the values of each time series as 
\begin{equation}
    \mathbf{X} = \{x_{1:L, 1:N}\} \in \mathbb{R}^{L \times N}
\end{equation}
where $L$ is the length of time series and $N$ is the number of features.
In this study, we allow each time series to have different lengths $L$, signifying personalized follow-up frequencies for different patient.
Consequently, we employ an observation mask $\mathbf{M}$ to represent missing values. $\mathbf{M}$ could be formulated as:
\begin{equation}
    \mathbf{M} = \{m_{1:L, 1:N}\} \in \{0, 1\}^{L \times N}
\end{equation}
where $m_{l,n}=0$ if $x_{l,n}$ is missing, and $m_{l,n}=1$ if $x_{l,n}$ is observed.

\paragraph{DNNs for EHR} 
For EHR data with missing values, traditional DNNs training protocol utilizes imputation matrices $\mathbf{I} = \{i_{1:L, 1:N}\} \in \mathbb{R}^{L \times N}$ to impute in the missing values. 
$\mathbf{I}$ is not readily available; instead, it needs to be generated based on imputation models.
The imputed EHR matrix $\mathbf{X}' \in \mathbb{R}^{L \times N}$ can be represented as:
\begin{equation}
    \mathbf{X}' = \{\mathbf{X}, \mathbf{I}\} = 
    \begin{cases}
        x_{l,n}, & \text{if } m_{l,n} \neq 0 \\
        i_{l,n}, & \text{if } m_{l,n} = 0
    \end{cases}
\end{equation}
For a plain DNNs, the imputed EHR data is embedded into a $d$-dimensional latent space by the backbone model of the DNNs:
\begin{align}
    \mathbf{e} &= \textcolor[RGB]{207,62,62}{\mathtt{backbone}}(\mathbf{X}') \\
    &= \textcolor[RGB]{207,62,62}{\mathtt{backbone}}(\{\mathbf{X}, \textcolor[RGB]{0,176,240}{\mathbf{I}}\}) \in \mathbb{R}^{d}
\end{align}

Since the $I$ is no longer updated during the DNNs training process, and to distinguish it further from the subsequent \TheName{} approach, we use the term \textcolor[RGB]{0,176,240}{frozen} for identification. Here, \textcolor[RGB]{207,62,62}{$\bullet$} and \textcolor[RGB]{0,176,240}{$\bullet$} represent the \textcolor[RGB]{207,62,62}{learnable} and \textcolor[RGB]{0,176,240}{frozen} parameters, respectively. 
Additionally, the backbone model in this context can be any type of multivariate time series processing model, including RNN, GRU \cite{cho2014learning}, Transformers \cite{vaswani2017attention}, or specialized models designed for EHR processing, such as ConCare \cite{ma2020concare}, SAFARI \cite{ma2022safari}, and others.

Furthermore, based on subsequent analysis tasks, DNNs employ specific head models to decode the vector $\mathbf{e}$ and make predictions. This process could be formulated as:
\begin{equation}
    \hat{\mathbf{y}} = \textcolor[RGB]{207,62,62}{\mathtt{head}}(\mathbf{e})
\end{equation}
Typically, DNNs use a Linear Layer for classification tasks and an Multi-Layer Perceptron (MLP) for regression tasks.

Finally, DNNs employ a loss function $Loss$ to perform gradient updates on all learnable parameters in the model. Typically, DNNs use Cross-Entropy loss for training classification tasks and Mean Squared Error (MSE) loss for training regression tasks.
\begin{equation}
    \mathcal{L}_{cls}(\mathbf{y}, \! \hat{\mathbf{y}}) \! = \! -\frac{1}{K} \! \sum_{i=1}^{K} \! \left( \mathbf{y}_i \! \cdot \! \log(\hat{\mathbf{y}}_i) \! + \! (1 \! -\!  \mathbf{y}_i) \! \cdot \! \log(1 \! - \! \hat{\mathbf{y}}_i) \right)\label{eq:celoss}
\end{equation}
\begin{equation}
    \mathcal{L}_{reg}(\mathbf{y}, \hat{\mathbf{y}}) = \frac{1}{K} \sum_{i=1}^{K} (\mathbf{y}_i - \hat{\mathbf{y}}_i)^2\label{eq:mseloss}
\end{equation}
where $\mathbf{y}$ and $\hat{\mathbf{y}}$ represent the true labels and predicted values, respectively. $K$ represents the number of samples in a mini-batch.

\subsection{Learnable Prompt as Pseudo-Imputation}\label{ssec:pai}
\TheName{} achieves direct learning of optimal pseudo-imputation for downstream tasks by introducing a feature prompt. During the training phase, \TheName{} updates the prompt based on the loss of the downstream task. In the inference phase, \TheName{} freezes the prompt and utilizes it for downstream clinical prediction tasks.

\paragraph{Feature prompt}
\TheName{} introduces a learnable prompt vector $\mathbf{V}$, and sets the length of $\mathbf{V}$ as $N$,
\begin{equation}
    \mathbf{V} = [v]_1, [v]_2, \dots, [v]_N \in \mathbb{R}^{N}
\end{equation}
which can be initialized arbitrarily, including random initialization or zero initialization. 
\TheName{} insert this prompt vector into the $\mathbf{X}$ based on $\mathbf{M}$, with different prompts inserted for different features.
\begin{equation}\label{eq:fill-prompt}
    \mathbf{X}' = \{\mathbf{X}, \textcolor[RGB]{207,62,62}{\mathbf{V}}\} = 
    \begin{cases}
        x_{l,n}, & \text{if } m_{l,n} \neq 0 \\
        \textcolor[RGB]{207,62,62}{v_{l,n}}, & \text{if } m_{l,n} = 0
    \end{cases}
\end{equation}

\paragraph{Training Phase}
\TheName{} inputs both $\mathbf{X}$ and $\mathbf{V}$ into DNNs, obtaining the $Loss$ through forward-propagation algorithm, and then updating DNNs and $\mathbf{V}$ using the back-propagation algorithm. This process can be formulated as:
\begin{equation}\label{eq:pai-back}
    \mathbf{e} = \textcolor[RGB]{207,62,62}{\mathtt{backbone}}(\{\mathbf{X}, \textcolor[RGB]{207,62,62}{\mathbf{V}}\})
\end{equation}
\begin{equation}\label{eq:pai-head}
    \hat{\mathbf{y}} = \textcolor[RGB]{207,62,62}{\mathtt{head}}(\mathbf{e})
\end{equation}

\TheName{} follows to the original training protocol, employing Cross-Entropy loss for training classification tasks (as shown in Eq.~\ref{eq:celoss}) and MSE loss for training regression tasks (as shown in Eq.~\ref{eq:mseloss}).

\paragraph{Asynchronous Learning}\label{sec:asy_learning}
Considering the heterogeneity between the structure of the $\mathtt{backbone}$\&$\mathtt{head}$ modules and the prompt $\mathbf{V}$, \TheName{} uses asynchronous learning to help DNNs better model the missing values while learning clinical tasks. 
Specifically, \TheName{} sets the parameters represented by the $\mathtt{backbone}$\&$\mathtt{head}$ and the $\mathbf{V}$ as $\theta_{model}$ and $\theta_{V}$ respectively, and then assigns different learning rates, $\eta_{model}$ and $\eta_{V}$, to them.

\TheName{} follows the standard training process to complete the forward propagation of EHR and prompt in the prediction model and calculate the loss $\mathcal{L}(\theta_{model}, \theta_{V})$, and then calculate the gradients for $\theta_{model}$ and $\theta_{V}$, respectively. 
\begin{equation}\label{eq:model-grad}
    g_{\theta_{model}} = \nabla_{\theta_{model}} \mathcal{L}(\theta_{model}, \theta_{V})
\end{equation}
\begin{equation}\label{eq:prompt-grad}
    g_{\theta_{V}} = \nabla_{\theta_{V}} \mathcal{L}(\theta_{model}, \theta_{V})
\end{equation}

\TheName{} uses parameter-specific learning rates to update the parameters.
\begin{equation}\label{eq:model-update}
    \theta_{model} = \theta_{model} - \eta_{model} \cdot g_{\theta_{model}}
\end{equation}
\begin{equation}\label{eq:prompt-update}
    \theta_{V} = \theta_{V} - \eta_{V} \cdot g_{\theta_{V}}
\end{equation}


\paragraph{Inference Phase}

Given DNNs and feature prompt $\mathbf{V}$ learned during the training phase, \TheName{} freezes DNNs and $\mathbf{V}$, then inputs $\mathbf{X}$ and $\mathbf{V}$ into DNNs, producing clinical prediction results for patients.

\begin{table*}[!ht]
\centering
\caption{Performance Comparisons for Mortality Outcome Prediction Task using Protocol of \textit{Impute-then-Regress Procedure} and \TheName{}. \textnormal{Best results among all methods are \textbf{bolded}. \textbf{mPS.}: min(P+,Se); \textbf{ROC.}: AUROC; \textbf{PRC.}: AUPRC. The experimental results of the t-test have been documented in the code repository\protect\footnotemark[1].}
}\label{tab:outcome-result}
\resizebox{1.0\textwidth}{!}{
\begin{tabular}{cccccccccccccc}
\toprule
\multirow{2}{*}{Models} & \multirow{2}{*}{Imputation} & \multicolumn{3}{c}{MIMIC-IV} & \multicolumn{3}{c}{CDSL}  & \multicolumn{3}{c}{Sepsis}  & \multicolumn{3}{c}{eICU} \\
\cmidrule(r){3-5} \cmidrule(r){6-8} \cmidrule(r){9-11} \cmidrule(r){12-14}
 & & \multicolumn{1}{c}{mPS.($\uparrow$)} & \multicolumn{1}{c}{ROC.($\uparrow$)} & \multicolumn{1}{c}{PRC.($\uparrow$)} & \multicolumn{1}{c}{mPS.($\uparrow$)} & \multicolumn{1}{c}{ROC.($\uparrow$)} & \multicolumn{1}{c}{PRC.($\uparrow$)} & \multicolumn{1}{c}{mPS.($\uparrow$)} & \multicolumn{1}{c}{ROC.($\uparrow$)} & \multicolumn{1}{c}{PRC.($\uparrow$)} & \multicolumn{1}{c}{mPS.($\uparrow$)} & \multicolumn{1}{c}{ROC.($\uparrow$)} & \multicolumn{1}{c}{PRC.($\uparrow$)} \\
\midrule
\multirow{3}{*}{RNN} 
& LOCF & 0.5033 & 0.8487 & 0.5177 & 0.8182 & 0.9637 & 0.8692 & 0.6133 & 0.9054 & 0.6870 & 0.4858 & 0.8227 & 0.5033 \\
& Zero & 0.4833 & 0.8124 & 0.5034 & 0.7818 & 0.9653 & 0.8678 & 0.6768 & 0.9252 & 0.7500 & 0.4364 & 0.7722 & 0.4318 \\
& \TheName{} & \textbf{0.5105} & \textbf{0.8494} & \textbf{0.5283} & \textbf{0.8889} & \textbf{0.9810} & \textbf{0.9322} & \textbf{0.6906} & \textbf{0.9450} & \textbf{0.7855} & \textbf{0.4966} & \textbf{0.8257} & \textbf{0.5089} \\
\multirow{3}{*}{LSTM} 
& LOCF & 0.5024 & 0.8517 & 0.5312 & 0.8148 & 0.9593 & 0.8593 & 0.6050 & 0.9064 & 0.6748 & 0.4786 & 0.8244 & 0.5068 \\
& Zero & 0.5065 & 0.8463 & 0.5099 & 0.8000 & 0.9273 & 0.8700 & 0.6466 & 0.9111 & 0.6981 & 0.4467 & 0.7794 & 0.4567 \\
& \TheName{} & \textbf{0.5327} & \textbf{0.8605} & \textbf{0.5364} & 0.8148 & \textbf{0.9815} & \textbf{0.8989} & \textbf{0.6951} & \textbf{0.9404} & \textbf{0.7767} & \textbf{0.5000} & \textbf{0.8331} & \textbf{0.5181} \\
\multirow{3}{*}{GRU} 
& LOCF & 0.5163 & 0.8631 & 0.5448 & 0.8148 & 0.9677 & 0.8862 & 0.5994 & 0.8950 & 0.6736 & 0.5292 & 0.8508 & 0.5580 \\
& Zero & 0.5068 & 0.8393 & 0.5193 & 0.8148 & 0.9766 & 0.9037 & 0.6630 & 0.9179 & 0.7341 & 0.4914 & 0.8145 & 0.5090 \\
& \TheName{} & \textbf{0.5212} & \textbf{0.8633} & \textbf{0.5472} & \textbf{0.8704} & \textbf{0.9800} & \textbf{0.9354} & \textbf{0.6685} & \textbf{0.9251} & \textbf{0.7507} & \textbf{0.5417} & \textbf{0.8557} & \textbf{0.5742} \\
\multirow{3}{*}{Transformer} 
& LOCF & 0.4780 & 0.8466 & 0.4665 & 0.7241 & 0.9492 & 0.7374 & 0.5691 & 0.8818 & 0.6165 & 0.4505 & 0.7928 & 0.4347 \\
& Zero & 0.4723 & 0.8287 & 0.4349 & 0.5926 & 0.9050 & 0.6626 & 0.5549 & 0.8700 & 0.6070 & 0.4450 & 0.7889 & 0.4309 \\
& \TheName{} & \textbf{0.4951} & \textbf{0.8467} & \textbf{0.4834} & \textbf{0.7368} & \textbf{0.9509} & \textbf{0.7670} & \textbf{0.6768} & \textbf{0.9299} & \textbf{0.7362} & \textbf{0.4838} & \textbf{0.8177} & \textbf{0.4786} \\
\midrule
\multirow{3}{*}{RETAIN} 
& LOCF & 0.5245 & 0.8643 & 0.5500 & 0.7963 & 0.9398 & 0.8166 & 0.5525 & 0.8816 & 0.6113 & 0.4492 & 0.7950 & 0.4395 \\
& Zero & 0.5204 & 0.8421 & 0.5056 & 0.7982 & 0.9700 & 0.8732 & 0.5773 & 0.8829 & 0.6324 & 0.4710 & 0.8012 & 0.4703 \\
& \TheName{} & \textbf{0.5539} & \textbf{0.8698} & \textbf{0.5597} & \textbf{0.8000} & \textbf{0.9800} & \textbf{0.8972} & \textbf{0.6116} & \textbf{0.9082} & \textbf{0.6716} & \textbf{0.4931} & \textbf{0.8213} & \textbf{0.5173} \\
\multirow{3}{*}{ConCare} 
& LOCF & 0.5229 & 0.8571 & 0.5403 & 0.7455 & 0.9443 & 0.8497 & 0.3996 & 0.7754 & 0.6937 & 0.5239 & 0.8478 & 0.5701 \\
& Zero & 0.5131 & 0.8451 & 0.5149 & 0.6481 & 0.9236 & 0.7544 & 0.4807 & 0.8477 & 0.7290 & 0.5233 & 0.8483 & 0.5715 \\
& \TheName{} & \textbf{0.5296} & \textbf{0.8630} & \textbf{0.5539} & \textbf{0.8545} & \textbf{0.9787} & \textbf{0.9090} & \textbf{0.7127} & \textbf{0.9360} & \textbf{0.7753} & \textbf{0.5420} & \textbf{0.8511} & \textbf{0.5736} \\
\multirow{3}{*}{M3Care} 
& LOCF & 0.5016 & 0.8453 & 0.5177 & 0.8333 & 0.9669 & 0.8633 & 0.6519 & 0.9008 & 0.7036 & 0.5111 & 0.8385 & 0.5428 \\
& Zero & 0.4886 & 0.8339 & 0.4885 & 0.7037 & 0.9332 & 0.7703 & 0.6492 & 0.9058 & 0.7155 & 0.4777 & 0.8155 & 0.5012 \\
& \TheName{} & \textbf{0.5131} & \textbf{0.8491} & \textbf{0.5300} & 0.8333 & \textbf{0.9762} & \textbf{0.9099} & \textbf{0.6796} & \textbf{0.9227} & \textbf{0.7560} & \textbf{0.5206} & \textbf{0.8500} & \textbf{0.5615} \\
\multirow{3}{*}{SAFARI} 
& LOCF & 0.5033 & 0.8582 & 0.5020 & 0.8148 & 0.9521 & 0.8566 & 0.6603 & 0.9218 & 0.7415 & 0.5481 & 0.8575 & 0.5826 \\
& Zero & 0.5095 & 0.8577 & 0.5132 & 0.8273 & 0.9495 & 0.8687 & 0.6339 & 0.9217 & 0.7106 & 0.5433 & 0.8540 & 0.5826 \\
& \TheName{} & \textbf{0.5154} & \textbf{0.8689} & \textbf{0.5305} & \textbf{0.9074} & \textbf{0.9828} & \textbf{0.9438} & \textbf{0.7210} & \textbf{0.9460} & \textbf{0.8042} & 0.5481 & \textbf{0.8608} & \textbf{0.5870} \\
\bottomrule
\end{tabular}
}
\end{table*}

\begin{table*}[!ht]
\centering
\caption{Performance Comparisons for Length-of-Stay Prediction Task using Protocol of \textit{Impute-then-Regress Procedure} and \TheName{}. \textnormal{Best results among all methods are \textbf{bolded}. The experimental results of the t-test have been documented in the code repository\protect\footnotemark[1].}
}\label{tab:los-result}
\begin{tabular}{ccccccccccc}
\toprule
\multirow{2}{*}{Models} & \multirow{2}{*}{Imputation} & \multicolumn{3}{c}{MIMIC-IV} & \multicolumn{3}{c}{CDSL} & \multicolumn{3}{c}{eICU} \\
\cmidrule(r){3-5} \cmidrule(r){6-8} \cmidrule(r){9-11}
 & & MSE ($\downarrow$) & RMSE ($\downarrow$) & MAE ($\downarrow$) & MSE ($\downarrow$) & RMSE ($\downarrow$) & MAE ($\downarrow$) & MSE ($\downarrow$) & RMSE ($\downarrow$) & MAE ($\downarrow$) \\
\midrule
\multirow{3}{*}{RNN} 
& LOCF & 2.9602 & 1.7205 & 1.0049 & 0.1458 & 0.3818 & 0.1580 & 1.6213 & 1.2733 & 0.9107 \\
& Zero & 2.9197 & 1.7087 & 0.9675 & 0.1609 & 0.4012 & 0.1638 & 1.6118 & 1.2696 & 0.9133 \\
& \TheName{} & \textbf{2.8383} & \textbf{1.6847} & \textbf{0.9580} & \textbf{0.1383} & \textbf{0.3720} & \textbf{0.1504} & \textbf{1.5520} & \textbf{1.2458} & \textbf{0.9037} \\
\multirow{3}{*}{LSTM} 
& LOCF & 2.9153 & 1.7074 & 0.9795 & 0.1434 & 0.3787 & 0.1730 & 1.6216 & 1.2734 & 0.8949 \\
& Zero & 2.9039 & 1.7041 & 0.9573 & 0.1324 & 0.3639 & 0.1495 & 1.5503 & 1.2498 & 0.9140 \\
& \TheName{} & \textbf{2.8747} & \textbf{1.6955} & \textbf{0.9460} & \textbf{0.1196} & \textbf{0.3458} & \textbf{0.1316} & \textbf{1.5403} & \textbf{1.2411} & \textbf{0.8908} \\
\multirow{3}{*}{GRU} 
& LOCF & 2.8944 & 1.7013 & 0.9860 & 0.1556 & 0.3945 & 0.1876 & 1.6024 & 1.2658 & 0.9028 \\
& Zero & 2.8958 & 1.7017 & 0.9746 & 0.1685 & 0.4105 & 0.1730 & 1.5672 & 1.2519 & 0.9232 \\
& \TheName{} & \textbf{2.8896} & \textbf{1.6998} & \textbf{0.9636} & \textbf{0.1303} & \textbf{0.3610} & \textbf{0.1694} & \textbf{1.4932} & \textbf{1.2220} & \textbf{0.9009} \\
\multirow{3}{*}{Transformer} 
& LOCF & 3.0805 & 1.7551 & 1.0347 & 0.1362 & 0.3691 & 0.1659 & 1.6196 & 1.2726 & 0.9157 \\
& Zero & 3.1220 & 1.7669 & 1.0197 & 0.1443 & 0.3799 & 0.1661 & 1.6186 & 1.2722 & 0.9228 \\
& \TheName{} & \textbf{3.0593} & \textbf{1.7491} & \textbf{1.0190} & \textbf{0.1253} & \textbf{0.3540} & \textbf{0.1345} & \textbf{1.5269} & \textbf{1.2357} & \textbf{0.9099} \\
\midrule
\multirow{3}{*}{RETAIN} 
& LOCF & 2.9668 & 1.7224 & 0.9784 & 0.1753 & 0.4187 & 0.2112 & 1.6361 & 1.2791 & 0.9127 \\
& Zero & 2.9872 & 1.7283 & 0.9464 & 0.1369 & 0.3699 & 0.1502 & 1.6051 & 1.2669 & 0.9262 \\
& \TheName{} & \textbf{2.8610} & \textbf{1.6915} & \textbf{0.9180} & \textbf{0.1259} & \textbf{0.3548} & \textbf{0.1362} & \textbf{1.4986} & \textbf{1.2242} & \textbf{0.9035} \\
\multirow{3}{*}{ConCare} 
& LOCF & 2.9278 & 1.7111 & 0.9858 & 0.1292 & 0.3595 & 0.1290 & 1.5977 & 1.2640 & 0.8962 \\
& Zero & 2.8916 & 1.7005 & 0.9819 & 0.1365 & 0.3656 & 0.1323 & 1.5434 & 1.2423 & 0.9110 \\
& \TheName{} & \textbf{2.8249} & \textbf{1.6807} & \textbf{0.9510} & \textbf{0.1282} & \textbf{0.3580} & \textbf{0.0879} & \textbf{1.4810} & \textbf{1.2170} & \textbf{0.8821} \\
\multirow{3}{*}{M3Care} 
& LOCF & 2.9616 & 1.7209 & 0.9956 & 0.1447 & 0.3804 & 0.1456 & 1.5976 & 1.2640 & 0.9006 \\
& Zero & 3.0013 & 1.7324 & 0.9876 & 0.1654 & 0.3979 & 0.1484 & 1.5952 & 1.2630 & 0.9224 \\
& \TheName{} & \textbf{2.9538} & \textbf{1.7187} & \textbf{0.9708} & \textbf{0.1288} & \textbf{0.3589} & \textbf{0.1100} & \textbf{1.5290} & \textbf{1.2365} & \textbf{0.8851} \\
\multirow{3}{*}{SAFARI} 
& LOCF & 2.9309 & 1.7120 & 0.9736 & 0.1744 & 0.4176 & 0.1741 & 1.5584 & 1.2483 & 0.8932 \\
& Zero & 2.8790 & 1.6967 & 0.9658 & 0.1527 & 0.3908 & 0.2128 & 1.5131 & 1.2301 & 0.9127 \\
& \TheName{} & \textbf{2.8031} & \textbf{1.6742} & \textbf{0.9604} & \textbf{0.1388} & \textbf{0.3726} & \textbf{0.1555} & \textbf{1.4306} & \textbf{1.1961} & \textbf{0.8871} \\
\bottomrule
\end{tabular}
\end{table*}

\section{Experiments}

\subsection{Dataset Descriptions}
We conduct experiments on four publicly available EHR datasets, including MIMIC-IV~\cite{johnson2023mimic}, CDSL~\cite{hm2020cdsl}, Sepsis~\cite{reyna2020early} and eICU~\cite{pollard2018eicu}, with two different types of prediction tasks (i.e., classification task and regression task) for evaluation. 
These datasets are real-world datasets widely used in existing research on DNNs for EHR \cite{gao2024comprehensive}.
The statistics of the four datasets are summarized in Appendix.~\ref{ssec:statistics-datasets}. 
The datasets are pre-processed following the pipeline in~\cite{gao2024comprehensive} with code~\cite{zhu2023pyehr}.

\subsection{Experimental Setups}

\paragraph{Implementation Details}
We employ a consistently uniform training and evaluation process to compare \TheName{} with traditional training protocols, including \textit{Impute-then-Regress Procedure} and \textit{Jointly Optimizing of Impute-n-Regress Procedure}, conducting comprehensive assessments with different DNNs and tasks. 
It is notable that in this experiment, we mainly compare \textit{Impute-then-Regress Procedure}. This is because existing \textit{Jointly Optimizing of Impute-n-Regress Procedure} has extremely strong limitations by itself (see Sec.~\ref{sec:Jo-I-a-R}). Moreover, our work focuses on better performance in clinical downstream tasks rather than imputation tasks, in this setting, \textit{Impute-then-Regress Procedure} has been widely used in recent related works on clinical prediction tasks~\cite{ma2020concare,ma2022safari,choi2016retain,zhang2022m3care}.
Further experimental implementation details and evaluation metrics can be found in Appendix.~\ref{ssec:implementation-details} and Appendix.~\ref{ssec:metrics}.

\paragraph{Baselines Models of Various EHR Analysis Tasks}

We conduct experiments with eight backbone models, including four general multivariate time series analysis models (RNN, LSTM, GRU \cite{cho2014learning} and Transformer \cite{vaswani2017attention}) and four specialized EHR analysis models (RETAIN \cite{choi2016retain}, ConCare \cite{ma2020concare}, M3Care \cite{zhang2022m3care} and SAFARI \cite{ma2022safari}). 
We use Linear and MLP as head models for different tasks, where Linear is used for classification tasks and MLP is used for regression tasks. 

\paragraph{Baselines Models of Various EHR Training Protocols}

For \textit{Impute-then-Regress Procedure}, we use Last Observation Carried Forward (LOCF)~\cite{wells2013strategies} and Zero value as the primary imputation model, which are widely used in various deep learning-based EHR analysis tasks \cite{ma2020concare,zhang2022m3care,ma2022safari}. 
Additionally, to further evaluate the performance of \TheName{}, we introduce recent DNN-based imputation models as baselines, including GPVAE \cite{fortuin2020gpvae}, CSDI \cite{tashiro2021csdi}, TimesNet \cite{wu2022timesnet} and SATIS \cite{du2023saits}. 
For \textit{Jointly Optimizing of Impute-n-Regress Procedure}, we introduce two works introduced in our related works as baseline models to verify the effectiveness of \TheName{}.

For more details, in the comparison between protocols of \textit{Impute-then-Regress Procedure} and \TheName{} for mortality outcome prediction and Length-of-Stay (LOS) prediction tasks, we employ the aforementioned eight backbone models. 
In subsequent analyses, we chose RNN and Transformer as representatives for further investigation. RNN and Transformer represent two distinct structural models widely applied in various EHR analysis tasks.

\begin{table*}[!ht]
\centering
\caption{Performance Comparisons for Mortality Outcome Prediction Task using Protocol of \textit{Impute-then-Regress Procedure} with Various Imputation Models and \TheName{}. \textnormal{Best results among all methods are \textbf{bolded}.}}\label{tab:imputation-result}
\begin{tabular}{cccccccc}
\toprule
\multirow{2}{*}{Models} & \multirow{2}{*}{Imputation} & \multicolumn{3}{c}{MIMIC-IV} & \multicolumn{3}{c}{CDSL} \\
\cmidrule(r){3-5} \cmidrule(r){6-8}
  & & \multicolumn{1}{c}{min(P+,Se)($\uparrow$)} & \multicolumn{1}{c}{AUROC($\uparrow$)} & \multicolumn{1}{c}{AUPRC($\uparrow$)} & \multicolumn{1}{c}{min(P+,Se)($\uparrow$)} & \multicolumn{1}{c}{AUROC($\uparrow$)} & \multicolumn{1}{c}{AUPRC($\uparrow$)} \\
 \midrule
\multirow{5}{*}{RNN} 
 & GPVAE & 0.4796 & 0.8356 & 0.4515 & 0.7818 & 0.9544 & 0.8262 \\
 & CSDI & 0.2720 & 0.6528 & 0.2251 & 0.6666 & 0.9171 & 0.7712 \\
 & TimesNet & 0.4869 & 0.8321 & 0.4683 & 0.8148 & 0.9609 & 0.8535 \\
 & SAITS & 0.4739 & 0.8383 & 0.4661 & 0.8182 & 0.9706 & 0.8884 \\
 & \TheName{} & \textbf{0.5105} & \textbf{0.8494} & \textbf{0.5283} & \textbf{0.8889} & \textbf{0.9810} & \textbf{0.9322} \\
 \midrule
\multirow{5}{*}{Transformer} 
 & GPVAE & 0.4738 & 0.8316 & 0.4361 & 0.7091 & 0.9455 & 0.7643 \\
 & CSDI & 0.3344 & 0.7267 & 0.3013 & 0.4815 & 0.8014 & 0.5048 \\
 & TimesNet & 0.4519 & 0.8298 & 0.4452 & 0.7037 & 0.9443 & 0.7102 \\
 & SAITS & 0.4421 & 0.8298 & 0.4166 & 0.6666 & 0.9446 & 0.7231 \\
 & \TheName{} & \textbf{0.4951} & \textbf{0.8467} & \textbf{0.4834} & \textbf{0.7368} & \textbf{0.9509} & \textbf{0.7670} \\
 \bottomrule
\end{tabular}
\end{table*}

\begin{table*}[!ht]
\centering
\caption{Performance Comparisons for Mortality Outcome Prediction Task using Protocol of \textit{Jointly Optimizing of Impute-n-Regress Procedure} and \TheName{}. \textnormal{Best results among all methods are \textbf{bolded}.}
}\label{tab:joint-optim}
\begin{tabular}{ccccccc}
\toprule
\multirow{2}{*}{Models} & \multicolumn{3}{c}{MIMIC-IV} & \multicolumn{3}{c}{CDSL} \\
\cmidrule(r){2-4} \cmidrule(r){5-7}
 & \multicolumn{1}{c}{min(P+,Se)($\uparrow$)} & \multicolumn{1}{c}{AUROC($\uparrow$)} & \multicolumn{1}{c}{AUPRC($\uparrow$)} & \multicolumn{1}{c}{min(P+,Se)($\uparrow$)} & \multicolumn{1}{c}{AUROC($\uparrow$)} & \multicolumn{1}{c}{AUPRC($\uparrow$)} \\
\midrule
SSGAN & 0.4232 & 0.7031 & 0.4873 & 0.7142 & 0.8366 & 0.7931 \\
MVIIL-GAN & 0.4563 & 0.7367 & 0.4792 & 0.8657 & 0.9013 & 0.8645 \\
RNN+\TheName{} & \textbf{0.5105} & \textbf{0.8494} & \textbf{0.5283} & \textbf{0.8889} & \textbf{0.9810} & \textbf{0.9322} \\
\bottomrule
\end{tabular}
\end{table*}

\subsection{Experimental Results}

\paragraph{Performance on various models and datasets}
Table.~\ref{tab:outcome-result} presents the performance of various EHR analysis models on mortality outcome prediction task of 4 datasets. 
In each configuration of DNNs, the performance between \textit{Impute-then-Regress Procedure} and the \TheName{} are compared.
Experimental results demonstrate that the \TheName{} outperforms the protocol of \textit{Impute-then-Regress Procedure} in terms of AUROC and AUPRC across all backbone models.
For the min(P+, Se) metric, \TheName{} achieves performance improvement in the majority of cases, with equivalent performance to the protocol of \textit{Impute-then-Regress Procedure} only observed when using LSTM and M3Care to analyze CDSL dataset, and using SAFARI to analyze eICU dataset.
Particularly noteworthy is the improvement in the crucial AUPRC metric, where \TheName{} achieves an average performance of \textcolor[RGB]{207,62,62}{1.06\%$\uparrow$} on MIMIC-IV, \textcolor[RGB]{207,62,62}{5.79\%$\uparrow$} on CDSL, \textcolor[RGB]{207,62,62}{8.18\%$\uparrow$} on Sepsis and \textcolor[RGB]{207,62,62}{2.27\%$\uparrow$} on eICU. This indicates that the \TheName{} protocol, without introducing non-authentic data, improves the performance of all existing EHR models.
Additionally, the results of the machine learning models can be found in the Appendix.~\ref{ssec:ml}.

\paragraph{Performance on various tasks}
Table.~\ref{tab:los-result} illustrates the performance of various EHR analysis models in predicting LOS on the MIMIC-IV, CDSL and eICU datasets. 
Experimental results indicate that in all cases, \TheName{} outperforms the protocol of \textit{Impute-then-Regress Procedure} comprehensively. Overall, compared to the protocol of \textit{Impute-then-Regress Procedure}, \TheName{} achieves an average performance of \textcolor[RGB]{207,62,62}{1.99\%$\uparrow$} in MSE, \textcolor[RGB]{207,62,62}{1.00\%$\uparrow$} in RMSE, and \textcolor[RGB]{207,62,62}{3.42\%$\uparrow$} in MAE on MIMIC-IV dataset. On CDSL dataset, it attains an average performance of \textcolor[RGB]{207,62,62}{12.16\%$\uparrow$} in MSE, \textcolor[RGB]{207,62,62}{6.19\%$\uparrow$} in RMSE, and \textcolor[RGB]{207,62,62}{22.51\%$\uparrow$} in MAE. This further underscores the performance advantage of \TheName{} on various tasks. Additionally, the results of the machine learning models can be found in the Appendix.~\ref{ssec:ml}.

\footnotetext[1]{\url{https://github.com/MrBlankness/PAI/blob/master/additional_results.md}}

\paragraph{Performance comparison of \TheName{} and more imputation models}
Table.~\ref{tab:imputation-result} displays the results of mortality outcome prediction task using various imputation models and \TheName{}. 
The results in indicate that \TheName{} outperforms all imputation models across all metrics. 
Additionally, we observed that under these imputation models, the performance of RNN and Transformer is inferior to that under LOCF imputation. 
For instance, with LOCF imputation, RNN achieved an AUPRC value of 51.77\%, while using SAITS as the imputation model, RNN obtained an AUPRC value of 45.15\% (\textcolor[RGB]{112,173,71}{6.62\%$\downarrow$}). 
Similar results were observed with other backbone models or datasets. 
This suggests that LOCF are superior, confirming the rationality of LOCF as a representative of imputation-based protocol.

\footnotetext[1]{\url{https://github.com/MrBlankness/PAI/blob/master/additional_results.md}}

\paragraph{Performance comparison of \TheName{} and jointly optimizing protocol}

Table.~\ref{tab:joint-optim} displays the results of mortality outcome prediction task using protocol of \textit{Jointly Optimizing of Impute-n-Regress Procedure} and \TheName{}.
It is worth noting that since both SSGAN and MVIIL-GAN complete the imputation task in a supervised manner, we randomly masked 20\% of the true values in the EHR data and used this as input to provide the supervision signal for the masked parts.
We analyze this experimental result from the following three aspects. 
\textbf{Performance}: We use the simplest RNN+\TheName{} as the representative of our method. The experimental results show that our method achieves the optimal result in the vast majority of cases. In the most important AUPRC metric, \TheName{} achieves a performance of \textcolor[RGB]{207,62,62}{4.10\%$\uparrow$} (on MIMIC-IV) and \textcolor[RGB]{207,62,62}{6.77\%$\uparrow$} (on CDSL) compared to the suboptimal model. 
\textbf{Parameter performance}: With settings of RNN on MIMIC-IV and CDSL, SSGAN introduces a GAN network (\textcolor[RGB]{112,173,71}{300.79\%$\uparrow$} and \textcolor[RGB]{112,173,71}{265.84\%$\uparrow$} parameters) to generate imputed data, while \TheName{} solely utilizes a learnable vector (\textcolor[RGB]{112,173,71}{1.98\%$\uparrow$} and \textcolor[RGB]{112,173,71}{2.31\%$\uparrow$} parameters) to model the downstream model's implicit preferences for missing values.
\textbf{Scalability:} \TheName{}, as an unsupervised method for imputation, far exceeds the semi-supervised SSGAN and the fully-supervised MVIIL-GAN in scalability.

\paragraph{Effectiveness of \textit{Asynchronous Learning}}

\begin{figure*}[!ht]
\centering
\includegraphics[width=0.98\textwidth]{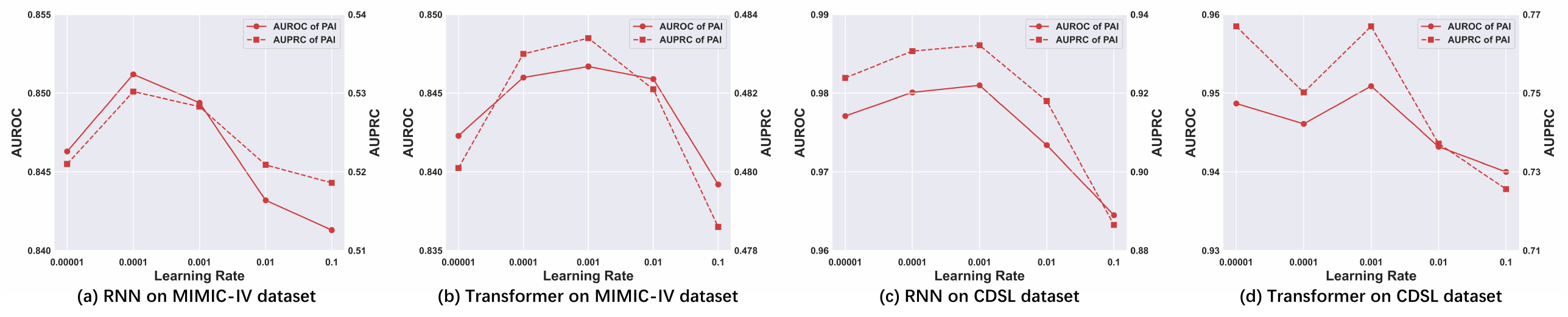}
\caption{Performance of \TheName{} with different learning rates on mortality outcome prediction task. The learning rate of the \texttt{backbone}\&\texttt{head} modules is fixed at 1e-2. Please note that AUROC and AUPRC are presented using a dual-axis plotting approach for better visualization.} \label{fig:lr}
\end{figure*}

To prove the effectiveness of asynchronous learning, we fixed the learning rates $\eta_{model}$ of the \texttt{backbone\&head} modules at 1e-2, and then set the learning rate $\eta_{V}$ of the prompt to different values, including 1e-1, 1e-2, 1e-3, 1e-4, and 1e-5, to evaluate the performance of \TheName{}. 
The experimental results are shown in Fig.~\ref{fig:lr}. 
The experimental results show that $\eta_{V} = 1e-3$ achieved the optimal performance in the vast majority of cases, only slightly inferior to $\eta_{V} = 1e-4$ in the setting of RNN on MIMIC-IV. 
Furthermore, we further observed that in all scenarios, \TheName{} performed better under the setting of $\eta_{V} < \eta_{model}$ than under the setting of $\eta_{V} \geq \eta_{model}$. 
Although the performance of \TheName{} was not stable under the setting of $\eta_{V} < \eta_{model}$, we still recommend using a value smaller than the $\eta_{model}$ of the main model as the $\eta_{V}$ of the prompt.


\begin{figure*}[!ht]
\centering
\includegraphics[width=0.98\textwidth]{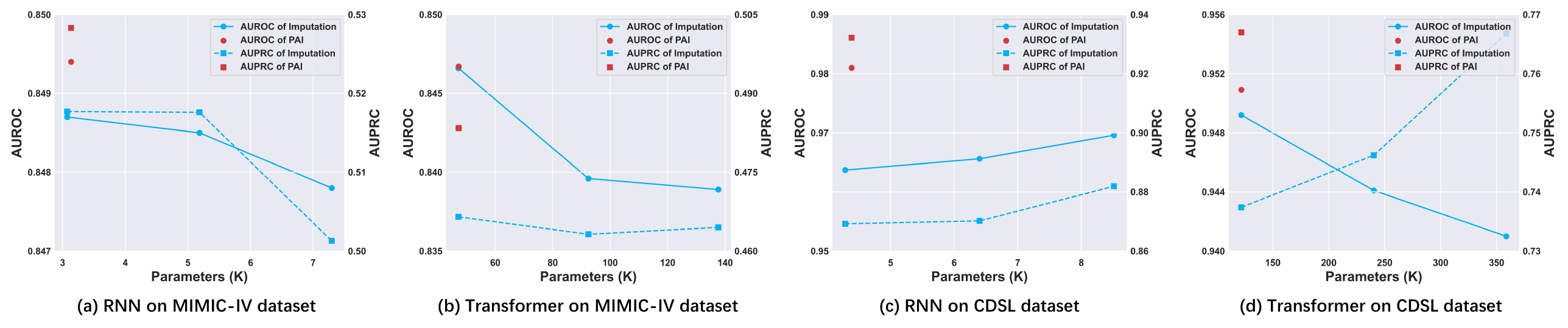}
\caption{Performance comparison of \TheName{} and imputation-based protocol with different parameters on mortality outcome prediction task. The imputation-based protocol uses RNN/Transformer with 1/2/3 layers respectively, and \TheName{} only use 1 layer. Please note that AUROC and AUPRC are presented using a dual-axis plotting approach for better visualization.} \label{fig:params}
\end{figure*}

\begin{figure*}[!ht]
\centering
\includegraphics[width=0.98\textwidth]{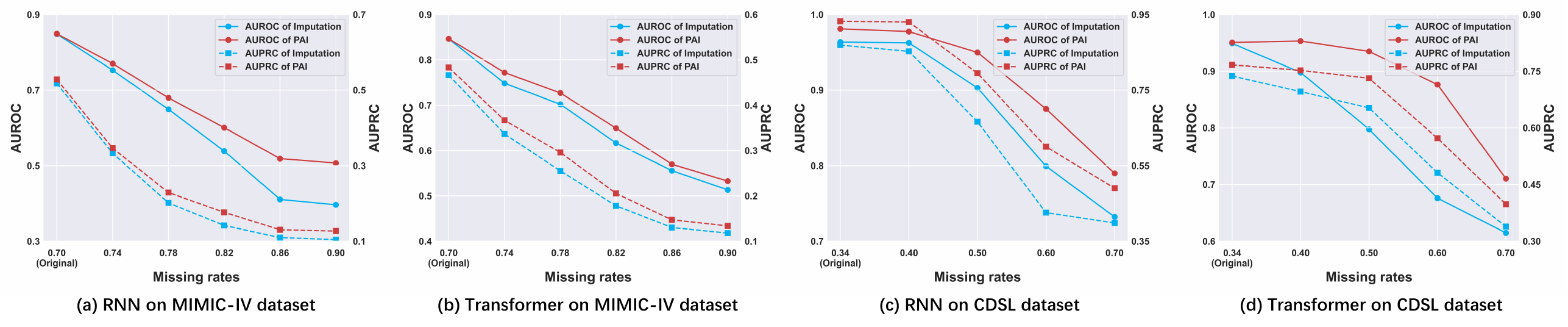}
\caption{Performance comparison of \TheName{} and imputation-based protocol with different missing rates on mortality outcome prediction task. Please note that AUROC and AUPRC are presented using a dual-axis plotting approach for better visualization.} \label{fig:miss-rate}
\end{figure*}

\section{Discussion}

In this section, we analyze the effectiveness of prompts. 
Although the inexplicability of prompts is still inferred in related works~\cite{zhou2022learning,zhou2022conditional}, we still present our insights through extensive experiments. 
To this end, we propose the following research questions:

\begin{itemize}
    \item \textbf{RQ\#1:} Does the benefit of prompts come from more parameters?
\end{itemize}

We have to admit that the introduction of \TheName{} adds additional learnable parameters, but they are very few. 
According to our statistics, among the 8 backbone models, \TheName{} only introduces \textcolor[RGB]{112,173,71}{1.98\%$\uparrow$} of the parameters in the task of MIMIC-IV and \textcolor[RGB]{112,173,71}{2.31\%$\uparrow$} of the parameters in CDSL. 
In addition, we further increase the number of parameters by introducing more layers on the imputation-based protocol to compare the performance of \TheName{}. The experimental results are shown in Fig.~\ref{fig:params}.
The experimental results show that there is no definite relationship between the increase in the number of parameters and the improvement in performance. 
For example, on the MIMIC-IV dataset, the increasing parameters of the model only leads to a decrease in the performance of the model. 
Although on the CDSL dataset, better performance can be achieved by increasing the parameters, it still does not exceed the results of \TheName{}. Therefore, we infer that the reason for the gain of \TheName{} comes from the direct learning of label information by the prompt rather than the introduction of more parameters.

\begin{itemize}
    \item \textbf{RQ\#2:} What does the prompt learns? Missing patterns or label information?
\end{itemize}

Some existing works~\cite{groenwold2020informative,tan2023informative}, suggest that there is a certain fixed pattern in data missing of EHR, that is, patients do not miss a certain examination for no reason. 
Since when using prompts, the model simultaneously learns information about the missing mask $\mathbf{M}$, therefore, a possible guess is that the prompt learns this missing pattern during this process and led to better performance.
We evaluated this hypothesis by randomly introducing more missing values. Detailed experimental settings can be found in the Appendix.~\ref{ssec:missing-rate}, and the experimental results are shown in Fig.~\ref{fig:miss-rate}.
Under this setting, the missing pattern in EHR was disrupted. Nevertheless, \TheName{} still stably surpassed the imputation-based protocol. 
Therefore, we infer that the effectiveness of \TheName{} is due to direct learning of current task labels rather than missing patterns.

\section{Conclusion}

In this work, we rethought the necessity of the imputation model in the traditional training protocol and proposed a new protocol, the Learnable Prompt as Pseudo-Imputation (\TheName{}), aimed at directly predict downstream tasks without the prior use of handcrafted or additional interpolation models for reconstructing missing values.
To verify the performance of \TheName{}, we conducted extensive experiments to validate this protocol. In the experiments, we first demonstrated that \TheName{} can function well on various datasets, various EHR analysis models, and various clinical prediction tasks, and comprehensively outperform the existing imputation-based protocols, then we validated the robustness of \TheName{} in analyses such as initialization strategies and  data insufficiency, and finally, we analyzed that the effectiveness of \TheName{} stems from direct learning of label information for downstream tasks rather than from additional parameters or implicit learning of missing patterns.
We claim that \TheName{} is a simple but highly scalable protocol that can be perfectly compatible with various EHR analysis tasks without introducing bias. In addition, \TheName{} allows for easy expansion in future work, and there remain many intriguing questions to explore. 
In summary, we hope the empirical results and insights presented in this work pave the way for future EHR analysis, inspiring the community to focus on modeling and analysis without imputation.

\begin{acks}

This work was supported by the National Natural Science Foundation of China (62402017, U23A20468), and Xuzhou Scientific Technological Projects (KC23143). 
Wen Tang acknowledges Clinical Cohort Construction Program of Peking University Third Hospital (No. BYSYDL2023004).
Liantao Ma acknowledges Peking University Medicine plus X Pilot Program-Key Technologies R\&D Project (2024YXXLHGG007).

\end{acks}

\bibliographystyle{ACM-Reference-Format}
\bibliography{main}

\appendix

\section{Details of Experimental settings}

\subsection{Statistics of the Datasets}\label{ssec:statistics-datasets}

\begin{table}[h]
\centering
\caption{\textit{Statistics of the MIMIC-IV, CDSL, Sepsis and eICU Datasets.}}
\label{tab:summary_statistics_dataset}
\resizebox{\columnwidth}{!}{
\begin{tabular}{lccc}
\toprule
  & Total & Alive & Dead \\
\midrule

\multicolumn{4}{c}{MIMIC-IV \cite{johnson2023mimic}} \\
\midrule
\# Patients &  56,888  & 51,451 (90.44 \%) & 5,437 (9.56 \%)  \\
\# Records & 4,055,519  & 3,466,575 (85.48 \%) & 588,944 (14.52 \%) \\
\# Avg. records & 42.0 [24.0, 75.0]  & 41.0 [24.0, 72.0] & 59.0 [24.0, 137.0]  \\
\midrule
\# Features & \multicolumn{3}{c}{61} \\
Length of stay & 42.1 [24.0, 75.2] & 41.0 [24.0, 71.9] & 61.2[25.2, 138.6] \\

\midrule

\multicolumn{4}{c}{CDSL \cite{hm2020cdsl}} \\
\midrule
\# Patients & 4,255 & 3,715 (87.31\%) & 540 (12.69\%) \\
\# Records  & 123,044 & 108,142 (87.89\%) & 14,902 (12.11\%) \\
\# Avg. records & 24.0 [15, 39] & 25.0 [15, 39] & 22.5 [11, 37] \\
\midrule
\# Features & \multicolumn{3}{c}{99} \\
Length of stay & 6.4 [4.0, 11.0] & 6.1 [4.0, 11.0] & 6.0 [3.0, 10.0]\\

\midrule

\multicolumn{4}{c}{Sepsis~\cite{reyna2020early}} \\
\midrule
\# Patients & 20.336 & 18,546 (91.20\%) & 1,790 (8.80\%) \\
\# Records  & 790,215 & 685,251 (86.72\%) & 104,964 (13.28\%) \\
\# Avg. records & 38.9 [8, 336] & 36.9 [8, 336] & 58.6 [8, 336] \\
\midrule
\# Features & \multicolumn{3}{c}{40} \\
Length of stay & N/A & N/A & N/A \\

\midrule

\multicolumn{4}{c}{eICU~\cite{pollard2018eicu}} \\
\midrule
\# Patients & 21,288 & 18,380 (86.34\%) & 2,908 (13.66\%) \\
\# Records  & 1,021,824 & 882,240 (86.34\%) & 139,584 (13.66\%) \\
\# Avg. records & 48.0 [48, 48] & 48.0 [48, 48] & 48.0 [48, 48] \\
\midrule
\# Features & \multicolumn{3}{c}{14} \\
Length of stay & 1.9 [0.0, 42.1] & 1.8 [0.0, 42.1] & 2.5 [0.0, 32.1]\\

\bottomrule
\end{tabular}
}
\end{table}

\subsection{Implementation Details} \label{ssec:implementation-details}
We implement our method using PyTorch v1.7.1 and conduct experiments on a machine equipped with an Nvidia Quadro RTX 8000 GPU. For each dataset, we employ stratified sampling to split the data into training, validation, and test sets in a 7:1:2 ratio. 
To ensure a fair comparison of different methods, we select model hyperparameters based on the performance on the validation set and report the performance on the test set. 
All methods are trained for 100 epochs using the Adam optimizer. 
We set the learning rate of DNNs to 1e-2, and for better prompt learning (see Sec.~\ref{sec:asy_learning}), we set the learning rate of prompts smaller than the learning rate of DNNs, eventually fix at 1e-3. 
No other learning rate decay or warm-up strategies are employed. We set the dimensionality of the latent representations for all models to 32.

\subsection{Evaluation Metrics}\label{ssec:metrics}
We assess the performance of the binary classification tasks with three evaluation metrics: area under the receiver operating characteristic curve (AUROC), area under the precision-recall curve (AUPRC), and the minimum of precision and sensitivity (Min(P+,Se)). 
Here we emphasize that among the three metrics, AUPRC is a more informative and primary metric when dealing with a highly imbalanced dataset like ours. 
For regression tasks, we select mean absolute error (MAE), mean squared error (MSE), and rooted-MSE (RMSE) as our evaluation metrics.

\begin{table*}[!ht]
\centering
\caption{Performance Comparisons for Mortality Outcome Prediction Task using Machine Learning Models. \textnormal{\textbf{mPS.}: min(P+,Se); \textbf{ROC.}: AUROC; \textbf{PRC.}: AUPRC.}
}\label{tab:outcome-result-ml}
\resizebox{1.0\textwidth}{!}{
\begin{tabular}{ccccccccccccc}
\toprule
\multirow{2}{*}{Models} & \multicolumn{3}{c}{MIMIC-IV} & \multicolumn{3}{c}{CDSL}  & \multicolumn{3}{c}{Sepsis}  & \multicolumn{3}{c}{eICU} \\
\cmidrule(r){2-4} \cmidrule(r){5-7} \cmidrule(r){8-10} \cmidrule(r){11-13}
 & \multicolumn{1}{c}{mPS.($\uparrow$)} & \multicolumn{1}{c}{ROC.($\uparrow$)} & \multicolumn{1}{c}{PRC.($\uparrow$)} & \multicolumn{1}{c}{mPS.($\uparrow$)} & \multicolumn{1}{c}{ROC.($\uparrow$)} & \multicolumn{1}{c}{PRC.($\uparrow$)} & \multicolumn{1}{c}{mPS.($\uparrow$)} & \multicolumn{1}{c}{ROC.($\uparrow$)} & \multicolumn{1}{c}{PRC.($\uparrow$)} & \multicolumn{1}{c}{mPS.($\uparrow$)} & \multicolumn{1}{c}{ROC.($\uparrow$)} & \multicolumn{1}{c}{PRC.($\uparrow$)} \\
\midrule
RF & 0.4461 & 0.8106 & 0.4344 & 0.7455 & 0.9414 & 0.7704 & 0.4282 & 0.8342 & 0.4231 & 0.4845 & 0.8204 & 0.4935 \\
DT & 0.4323 & 0.8094 & 0.3986 & 0.6667 & 0.9233 & 0.6415 & 0.4365 & 0.8169 & 0.3990 & 0.4384 & 0.7944 & 0.4207 \\
GBDT & 0.4708 & 0.8319 & 0.4619 & 0.7273 & 0.9504 & 0.7821 & 0.4862 & 0.8655 & 0.4083 & 0.4974 & 0.8337 & 0.4977 \\
CatBoost & 0.4667 & 0.8254 & 0.4479 & 0.7407 & 0.9535 & 0.8015 & 0.4669 & 0.8609 & 0.4890 & 0.4897 & 0.8271 & 0.5041 \\
XGBoost & 0.4690 & 0.8327 & 0.4596 & 0.7593 & 0.9558 & 0.7950 & 0.5262 & 0.8763 & 0.5047 & 0.4991 & 0.8304 & 0.5049  \\
\bottomrule
\end{tabular}
}
\end{table*}

\begin{table*}[!ht]
\centering
\caption{Performance Comparisons for Length-of-Stay Prediction Task using Machine Learning Models.}
\label{tab:los-result-ml}
\begin{tabular}{cccccccccc}
\toprule
\multirow{2}{*}{Models} & \multicolumn{3}{c}{MIMIC-IV} & \multicolumn{3}{c}{CDSL} & \multicolumn{3}{c}{eICU} \\
\cmidrule(r){2-4} \cmidrule(r){5-7} \cmidrule(r){8-10}
 & MSE ($\downarrow$) & RMSE ($\downarrow$) & MAE ($\downarrow$) & MSE ($\downarrow$) & RMSE ($\downarrow$) & MAE ($\downarrow$) & MSE ($\downarrow$) & RMSE ($\downarrow$) & MAE ($\downarrow$) \\
\midrule
RF & 3.2519 & 1.8033 & 1.1969 & 0.1718 & 0.4068 & 0.1995 & 2.3056 & 1.5184 & 1.2786 \\
DT & 3.2941 & 1.8150 & 1.2005 & 0.1749 & 0.4190 & 0.2016 & 2.3535 & 1.5341 & 1.2835 \\
GBDT & 3.2337 & 1.7982 & 1.1959 & 0.1713 & 0.4096 & 0.1987 & 2.3206 & 1.5233 & 1.2812 \\
CatBoost & 3.2387 & 1.7996 & 1.1993 & 0.1791 & 0.3994 & 0.1930 & 2.3035 & 1.5177 & 1.2813 \\
XGBoost & 3.2305 & 1.7974 & 1.1942 & 0.1742 & 0.4043 & 0.1989 & 2.3103 & 1.5200 & 1.2789 \\
\bottomrule
\end{tabular}
\end{table*}

\section{Additional Experimental Results}

\subsection{Performance of Two Task using Machine Learning Models} \label{ssec:ml}

To provide a comprehensive benchmark comparison between existing models, we further introduce machine learning models for baseline comparison, including Decision tree (DT), Random forest (RF), Gradient Boosting Decision Tree (GBDT), XGBoost and CatBoost.

Table.~\ref{tab:outcome-result-ml} and Table.~\ref{tab:los-result-ml} present the performance of machine learning models on the mortality outcome and length-of-stay prediction task, indicating that the performance of all machine learning models is generally lower than that of deep learning models. 
As \TheName{} optimizes the prompt based on the loss function, in this work, we do not use \TheName{} to enhance the performance of machine learning models on EHR with missing values.



\begin{figure*}[t]
\centering
\includegraphics[width=0.98\textwidth]{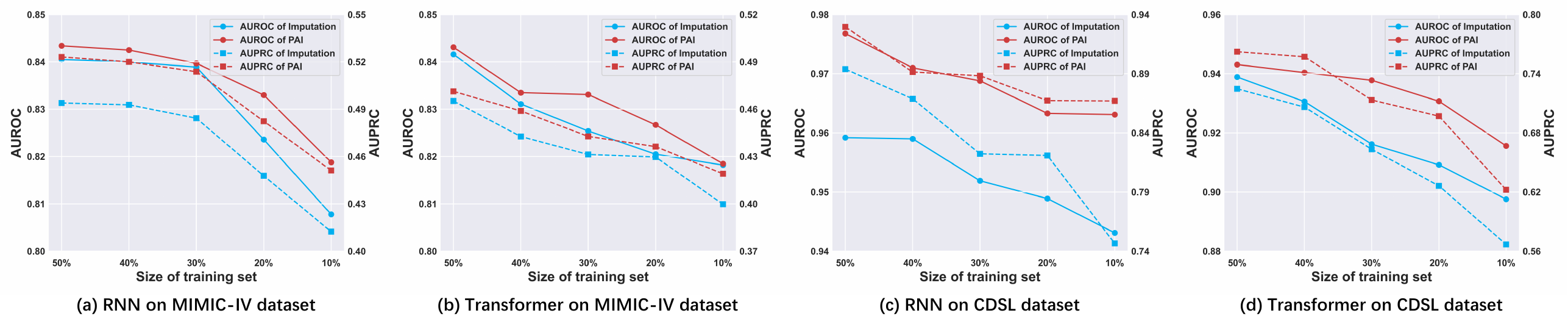}
\caption{Performance comparison of \TheName{} and imputation-based protocol with different amount of training data on mortality outcome prediction task. Please note that AUROC and AUPRC are presented using a dual-axis plotting approach for better visualization.} \label{fig:sample-rate}
\end{figure*}

\subsection{Robustness against data insufficiency}~\label{ssec:data-sample}

We evaluate the robustness of \TheName{} against insufficient training samples.  
We reduce the sample size (i.e., the number of patients) of the training set from 70\% to 50\%/40\%/30\%/20\%/10\% of the whole dataset to simulate the scenario of data insufficiency, and the test set is fixed for a fair comparison. We conduct experiments on those different settings, and the performances of AUROC and AUPRC are plotted in Fig.~\ref{fig:sample-rate}. 
As is shown in Fig.~\ref{fig:sample-rate}, \TheName{} consistently outperforms all selected baselines under all settings. 
Furthermore, as the size of the training set shrinks, the performances of imputation-base protocol decrease more sharply than ours, leading to a more significant performance gap. 
For example, in the case of RNN on the CDSL dataset, even when we adopt only 10\% of data for training, \TheName{} still reaches an AUROC of 96.31\% and an AUPRC of 86.70\%, while imputation-base protocol drop to 94.31\% AUROC (\textcolor[RGB]{112,173,71}{2.00\%$\downarrow$}) and 74.66\% AUPRC (\textcolor[RGB]{112,173,71}{12.04\%$\downarrow$}). Those results indicate that \TheName{} is more tolerant of data insufficiency with greater robustness.

\subsection{Robustness against higher missing rates}~\label{ssec:missing-rate}
We evaluate the robustness of \TheName{} on data with higher missing rates.
We randomly drop values in the training/validation/test set, adjust the missing rate of MIMIC-IV from 70\% to 74\%/78\%/82\%/86\%/90\%, and the missing rate of CDSL from 34\% to 40\%/50\%/60\%/70\% to simulate a higher proportion of missing values.
To ensure the same data distribution, we apply the same operations to the validation and test sets. 
We ensure that manually created missing values are exactly consistent when comparing with baseline models. 
The performance in terms of AUROC and AUPRC is plotted in Fig.~\ref{fig:miss-rate}. 
From it we observe a sharp decline in performance as the missing values gradually increase, for both \TheName{} and imputation-based protocol. 
However, similar to the results with insufficient data, \TheName{} exhibits higher robustness compared to imputation-based protocol. 
For instance, in the case of RNN on the MIMIC dataset, when the missing rate is the original 70\%, \TheName{} achieves a 0.07\%$\uparrow$ AUROC and 1.06\%$\uparrow$ AUPRC compared to imputation-based protocol, and when the missing rate increases to 90\%, \TheName{} achieves 11.03\%$\uparrow$ (\textcolor[RGB]{207,62,62}{10.96\%$\uparrow$}) AUROC and 2.27\%$\uparrow$ (\textcolor[RGB]{207,62,62}{1.21\%$\uparrow$}) AUPRC.

\subsection{Evaluations on More Benchmarks}

We further evaluate the performance of \TheName{} on additional benchmarks, including the Mortality Outcome and Length-of-Stay Prediction Task on the MIMIC-III dataset, as well as the Readmission Task on the MIMIC-IV dataset. The experimental results are documented in the code repository\footnote{\url{https://github.com/MrBlankness/PAI/blob/master/additional_results.md}}. The results demonstrate that \TheName{} maintains sufficient competitiveness compared to traditional imputation-based protocols across these additional benchmarks.

\end{document}